\documentclass[times,twocolumn,final,authoryear]{elsarticle}

\usepackage{prletters}
\usepackage{framed,multirow}

\setcitestyle{numbers,square}

\usepackage[labelformat=simple]{subcaption}
\DeclareCaptionLabelSeparator{periodspace}{.\;}
  \usepackage{grffile}
\usepackage{afterpage}
\usepackage{booktabs}

\usepackage{amsmath, amssymb, amsfonts, mathtools}
\usepackage{bm,dsfont}
\usepackage{latexsym}
\usepackage{kbordermatrix}

\usepackage{framed,multirow}

\usepackage{todonotes}

\usepackage{url}
\usepackage{xcolor}
\definecolor{newcolor}{rgb}{.8,.349,.1}

\newcommand{\R}{{\mathbb R}}

\newcommand{\E}{{\mathbb E}}
\newcommand{\T}{{\mathcal{T}}}

\newcommand{\X}{{\mathcal{X}}}

\newcommand{\norm}[1]{\left\lVert#1\right\rVert}

\renewcommand{\S}{{\mathcal{S}}}

\DeclareMathOperator*{\sign}{sign}

\usepackage[normalem]{ulem}

\journal{Pattern Recognition Letters}

\usepackage[colorlinks=false]{hyperref}

\def\itemautorefname~#1\null{#1\null}

\begin{document}

\setlength{\abovedisplayskip}{5pt}
\setlength{\belowdisplayskip}{4pt}

\thispagestyle{empty}

\begin{frontmatter}

\title{Generating Labels for Regression of Subjective Constructs using Triplet Embeddings}
\tnotetext[t1]{\url{https://doi.org/10.1016/j.patrec.2019.10.003}}

\author[1]{Karel \snm{Mundnich}\corref{cor1}}
\cortext[cor1]{Corresponding author:
  Tel.: +1-213-740-4146;
  fax: +1-213-740-4651;}
\ead{mundnich@usc.edu}
\author[1]{Brandon M. \snm{Booth}}
\author[1]{Benjamin \snm{Girault}}
\author[1]{Shrikanth \snm{Narayanan}}

\address[1]{Signal Analysis and Interpretation Lab, University of Southern California, 3740 McClintock Ave EEB 400, Los Angeles CA 90089, USA}

\received{1 May 2013}
\finalform{10 May 2013}
\accepted{13 May 2013}
\availableonline{15 May 2013}
\communicated{S. Sarkar}

\begin{abstract}
Human annotations serve an important role in computational models where the target constructs under study are hidden, such as dimensions of affect. This is especially relevant in machine learning, where subjective labels derived from related observable signals (e.g., audio, video, text) are needed to support model training and testing. Current research trends focus on correcting artifacts and biases introduced by annotators during the annotation process while fusing them into a single annotation. In this work, we propose a novel annotation approach using triplet embeddings. By replacing the absolute annotation process to relative annotations where the annotator compares individual target constructs in triplets, we leverage the accuracy of comparisons over absolute ratings by human annotators. We then build a 1-dimensional embedding in Euclidean space that is indexed in time and serves as a label for regression. In this setting, the annotation fusion occurs naturally as a union of sets of sampled triplet comparisons among different annotators.
We show that by using our proposed sampling method to find an embedding, we are able to accurately represent synthetic hidden constructs in time under noisy sampling conditions. We further validate this approach using human annotations collected from Mechanical Turk and show that we can recover the underlying structure of the hidden construct up to bias and scaling factors.
\end{abstract}


\end{frontmatter}

\section{Introduction}\label{sec:introduction}
Continuous-time annotations are an essential resource for the computational study of hidden constructs such as human affect or behavioral traits over time. Indeed, the study of these hidden constructs is commonly tackled using regression techniques under a supervised learning framework, which heavily rely on accurately labeled features with respect to the constructs under study. Formally, regression problems deal with finding a mapping $f: \mathcal{X} \to \mathcal{Y}$, where $\mathcal{X}$ is the \emph{feature space}, and $\mathcal{Y}$ is the \emph{label space}. Note that if $\bm{Y}\in\mathcal{Y}$ is indexed by time, then it is sometimes called a \emph{continuous-time label}\footnote{As opposed to discrete labels without time dependency.}. In this paper, we are interested in finding labels $\bm{Y}\in\mathcal{Y}$, such that $\bm{Y}$ is a good proxy for a hidden construct $\bm{Z}\in\mathcal{Z}$. As an example, in affective computing, $\bm{Z}$ is often a dimension of affect such as arousal (emotion intensity) or valence (emotion polarity), and it is assumed to be characterizable by data in the observation space $\mathcal{X}$ (e.g. audio, video, or bio-behavioral signals).

In the current literature, continuous-time labels in $\mathcal{Y} \subseteq \R^n$ are often generated from a set of continuous-time annotations acquired from a set of human raters or annotators $\mathcal{A}$. Each annotator $a\in\mathcal{A}$ uses perceptually interpretable features $\bm{X}\in\mathcal{X}\subseteq\R^{n\times q}$ to generate annotations $\bm{Y}_a\in\mathcal{Y}_a\subseteq\R^n$ about the construct $\bm{Z}$ \cite{Cowie2003, Metallinou2013, Busso2013}. In the sets above, $n$ is the number of samples in time, and $q$ represents the dimension of the set of perceptual features (e.g. audio levels, frames in a video) used for the real-time annotation acquisition. More generally, annotators are requested to do the mapping:
\begin{align}
    f^a_{\bm{Z}}: \mathcal{X} &\to \mathcal{Y}_a, \\
    \bm{X} &\mapsto f^a_{\bm{Z}}(\bm{X}) = \bm{Y}_a,
\end{align}
where each $f^a_{\bm{Z}}$ is specific to annotator $a$ for a construct $\bm{Z}$.
Usually, several of these single annotations $\bm Y_a$ are collected from several annotators $a\in\mathcal{A}$, processed, and combined to create a single label $\bm{Y}$. This problem is called \emph{annotation fusion}.

To train accurate statistical models, it is important that the labels $\bm{Y}$ used are precise and accurate, and properly reflect the variable $\bm{Z}$ under study \cite{Raykar2010}. Unfortunately, the annotation of hidden cues such as behavioral traits is a challenging problem due to several factors including diverse interpretations of the construct under study, differences in the perception of scale, improper design of the annotation-capturing tools, as well as disparate reaction times \citep{Metallinou2013, Mariooryad2013, Booth2018}. All of these affect the fidelity of individual annotations $\bm{Y}_a$.

\begin{figure}[t!]
    \centering
    \definecolor{mycolor1}{rgb}{0.121568627451,0.466666666667,0.705882352941}
    \definecolor{mycolor2}{rgb}{1.0,0.498039215686,0.0549019607843}
    \definecolor{mycolor3}{rgb}{0.172549019608,0.627450980392,0.172549019608}
    \definecolor{mycolor4}{rgb}{0.839215686275,0.152941176471,0.156862745098}
    \definecolor{mycolor5}{rgb}{0.580392156863,0.403921568627,0.741176470588}
    \definecolor{mycolor6}{rgb}{0.0901960784314,0.745098039216,0.811764705882}
    \begin{subfigure}{\linewidth}
        \centering
        \includegraphics{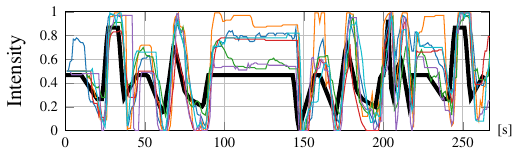}
        \caption{Task A: \url{https://youtu.be/hOPfInpDD9E}}
        \label{fig:example_TaskA}
    \end{subfigure}
    \begin{subfigure}{\linewidth}
       \centering
       \includegraphics{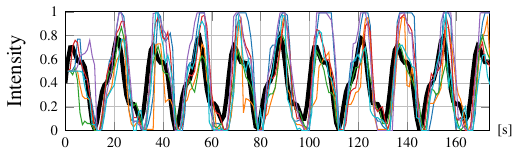}
       \caption{Task B: \url{https://youtu.be/o3o5cUPBPAg}}
        \label{fig:example_TaskB}
    \end{subfigure}
    \caption{
    Two real-time human annotation tasks with known ground truth $\bm{Z}$ (intensity of green over time, shown by the thick black lines). The annotators were presented with a user interface in which the video was shown, and they had to move a slider to match in real-time the current intensity of green observed. Six annotations are plotted in each task. Different colors represent each annotation $\bm{Y}_a$ done in real-time by a different annotator $a\in\mathcal{A}$ in a synthetic data experiment. For more details, please refer to \cite{Booth2018}.
    }
    \label{fig:example}
    \vspace{-0.4cm}
\end{figure}

To better study these challenges and the efficacy of algorithms to generate $\bm{Y}$, we build upon perceptual annotation tasks proposed previously in \citep{Booth2018} where the ground truth $\bm{Z}$ is known, as a way to evaluate annotation fusion and correction algorithms. We proposed these tasks to decouple the problems of annotations themselves and the interpretation of hidden constructs.
\autoref{fig:example} shows the outcome of these experiments in \cite{Booth2018}, where nine human annotators were asked to annotate the intensity of green color (varying continuously between 0 and 1) in two different tasks (A and B) by moving a slider while watching the videos to match the intensity they were observing. We invite the readers to look at the videos referenced in the caption of \autoref{fig:example} to directly experience what was presented to the annotators.
More complex real-world scenarios with coupled problems will be the subject of a future communication.
In \autoref{fig:example}, six annotations are plotted for clarity for each task. \autoref{fig:example} exhibits many of the artifacts that complicate the fusion of continuous-time annotations: variable reaction times \cite{Mariooryad2015}, overshooting fast changes, time-varying biases, disparate interpretations of scale, and difficulties in annotating constant intervals of the variable under study (mainly due to real-time corrections in the annotation process of the annotators themselves).

\subsection{Related work} \label{sec:introduction:literature}
Related recent research has attempted to estimate an underlying construct $\bm{Z}$ by using continuous-time annotations $\bm{Y}_a$. Different works have addressed a subset of the aforementioned challenges (time lags, scale interpretations). For example, \citep{Mariooryad2013, Mariooryad2015} study and model the reaction lag of annotators by using features from the data and shift each annotation before performing a simple average to fuse them, thus creating a unique label (EvalDep). Dynamic time warping (DTW) proposed by \citep{DTW2007} is another popular time-alignment method that warps the signals in time to maximize time-alignment, which is usually combined with weighted averaging of signals. \citep{Ringeval2015} proposes the use of a Long-Short-Term-Memory network (LSTM) to fuse asynchronous input annotations, by conducting time-alignment and de-biasing the different annotations. \citep{gupta2016modeling} presents a method for modeling multiple annotations over a continuous variable, and computes the ground truth by modeling annotator-specific distortions as filters whose parameters can be estimated jointly using Expectation-Maximization (EM). However, this work relies on heavy assumptions in the models for mathematical tractability, that do not necessarily reflect how annotators behave. All of the aforementioned works involve post-processing the raw continuous-time annotations, and performing the annotation fusion by averaging weighted signals in different (non)linear ways.

A different set of approaches is used to learn a warping function so that the fusion better correlates with associated features \citep{Hotelling1936, Nicolaou2013}. These spatial-warping methods can be combined with time warping \citep{Zhou2009, zhou2012generalized, trigeorgis2018deep}. All of these approaches rely on using a set of features.

In \citep{Booth2018}, we proposed a framework based on triplet embeddings to correct a continuous-time label generated by a fusion algorithm. This approach warps the fused label by selecting specific windows of it in time to collect extra information from human annotators through triplet comparisons. In \citep{Booth2018AVEC}, we also used triplet embeddings to fuse real-time annotations directly, by using majority voting to make a decision for each query. However, in these works the question of whether triplet comparisons \textit{alone} can be used to generate the label $\bm{Y}$ is not studied.

A Triplet Embedding approach to learn metrics from multi-modal data was first proposed in \citep{mcfee2011learning}. The authors develop an algorithm to account for noisy triplet labels (the notion of noisy labels was initially observed by \cite{ellis2002quest} in music applications). In \citep{mcfee2011learning}, the authors use their proposed algorithm to embed artists based on their (subjective) similarities.
In \cite{mcfee2012learning} the authors introduce the idea of using ranking information extracted from metric leaning approaches for the comparison of music applied to recommender systems.
However, none of these works use triplet embeddings to model the dynamics of subjective constructs over time. This is the topic of this paper.

\subsection{Contributions}
In this paper we study the performance of a new methodology to acquire and create a single label for regression by changing the sampling procedure of the latent construct. We sample this information by asking annotators questions of the form ``is the signal in time-frame $i$ more similar to the signal in time-frame $j$ or $k$?'' to build a 1-dimensional embedding $\bm{Y}$ in Euclidean space, where $(i,j,k)$ forms a \emph{triplet}. \autoref{fig:MTurk_question_design} shows an example of a query in the proposed sampling method where the comparison is based on the perceived shade (intensity) of the color.

Formally, we propose that annotators perform the following mapping:
\begin{align}
    f_\mathcal{Z}^a : \bm{\X} \times \bm{\X} \times \bm{\X} &\to \{-1,+1\},\\
    (\bm{x}_i, \bm{x}_j, \bm{x}_k) &\mapsto \sign\left(d_\mathcal{Z}^a(\bm{x}_i,\bm{x}_k) - d_\mathcal{Z}^a(\bm{x}_i,\bm{x}_j)\right) = w_t^a,
\end{align}
where $d_{\smash{\mathcal{Z}}}^a$ is the perceived dissimilarity of construct $\mathcal{Z}$ by annotator $a$. We use a set of queried triplets $\{(i,j,k)\}$ and the corresponding annotations $\{w^a_t = \sign(d_{\smash{\mathcal{Z}}}^a(\bm{x}_i,\bm{x}_k) - d_{\smash{\mathcal{Z}}}^a(\bm{x}_i,\bm{x}_j))\}$ to calculate the embedding $\bm{Y}$.

We motivate this approach using three key observations. First, psychology and machine learning/signal processing studies have shown that people are better at comparing than rating items \citep{Stewart2005, Yannakakis2011, Metallinou2013, Yannakakis2015}, so this sampling mechanism is easier for annotators than requesting absolute ratings in real-time. Second, the use of triplet embeddings naturally solves the annotation fusion problem, since it is done by taking the union of sets (details in \autoref{sec:approach}). Third, triplet embeddings offer a simple way of verifying the agreement of the annotations, given by the number of triplet violations in the computed embedding.

We empirically show that it is possible to reconstruct the hidden green intensity signal (i.e., recover the metric information) of tasks A and B in \autoref{fig:example} under different synthetic noise scenarios in the triplet labeling stage. These reconstructions are accurate up to a scaling and bias factor but do not suffer from artifacts such as time-lags present in real-time annotations. Moreover, to test our approach, we gather triplet comparisons for the same experiments from human annotators in Amazon Mechanical Turk and show that it is possible to reconstruct the hidden green intensity values over time up to scaling and bias factors when humans perform the triplet comparisons. Finally, we compare our results to two continuous-annotation fusion algorithms recently proposed in the literature to show the strengths of our method.

\section{Background: Triplet Embeddings}\label{sec:background}
We first recall the general setting of Triplet Embeddings from a probabilistic perspective \cite{Jain2016}. Let $\bm{z}_1,\ldots,\bm{z}_n$ be items that we want to represent through points $\bm{y}_1,\ldots,\bm{y}_n\in\R^{m}$, respectively, with $[\bm{y}_1\ldots\bm{y}_n] = \bm{Y}\in\R^{m\times n}$. We assume that the items $\{\bm{z}_i\}$ lie in a metric space, and the Euclidean distances between them are given by $\bm{D}^*_{ij} = \norm{\bm{z}_i - \bm{z}_j}^2_2$. We also assume that we have access to noisy distance comparisons, denoted by $d(\bm{z}_i, \bm{z}_j)$. These noisy distances may be perceptual, such as comparisons of expressed affect in the context of affective computing. We use these noisy distances to examine comparisons of the form:
\begin{equation}
d(\bm{z}_i, \bm{z}_j) \stackrel{?}{\lessgtr} d(\bm{z}_i, \bm{z}_k)
\label{eq:decision}
\end{equation}
to find the embedding $\bm{Y}$.

Formally, let $\mathcal{T}$ be the set of all possible unique triplets for $n$ items:
\begin{equation}
    \mathcal{T} = \{(i,j,k) \;|\; i \neq j
    < k \neq i, 1 \leq i,j,k \leq n \}.
\end{equation}
Note that $|\mathcal{T}| = n\binom{n-1}{2} = \mathcal{O}(n^3)$, which may be a very large set. We \textit{observe} a set of triplets $\S$, such that $\S \subseteq \T$, and corresponding realizations of the random variables $w_t$, where $t = (i,j,k)\in\S$, such that:
\begin{equation}
    w_t =
    \begin{cases}
      -1, &\text{w.p. } f(\bm{D}^*_{ij} - \bm{D}^*_{ik})\\
      +1, &\text{w.p. } 1 - f(\bm{D}^*_{ij} - \bm{D}^*_{ik}).
    \end{cases}\label{eq:triplet_noise}
\end{equation}
Here, $f:\R\to[0,1]$ is a function that behaves as a cumulative distribution function \citep{davenport2014} (sometimes called link function), and therefore has the property that $f(-x) = 1 - f(x)$. Hence, the $w_t$'s indicate if $i$ is closer to $j$ than $k$, with a probability depending on the difference $\bm{D}^*_{ij} - \bm{D}^*_{ik}$ (or the difficulty of the annotation task).

Let $\bm{G} = \bm{Y}^\top\bm{Y}$ be the Gram matrix of the embedding. We can estimate $\bm{G}$ (and hence $\bm{Y}$) by minimizing the empirical risk:
\begin{equation}\label{eq:empirical_risk}
    \widehat{R}_\S (\bm{G}) = \frac{1}{|\S|}\sum_{t\in\S} \ell\left( w_t \langle \bm{\mathcal{L}}_t, \bm{G}\rangle_F\right),
\end{equation}
where $\ell$ is a (margin-based) loss function and $\bm{\mathcal{L}}_t$ is defined as:
\def\shortminus{\scalebox{0.5}[1.0]{$-$}}
\def\ps{\phantom{\shortminus}}
\begin{equation}
    \bm{\mathcal{L}}_t = \kbordermatrix{
        &  i  &  j  &  k  \\
      i &  \ps 0  & \shortminus1  &  \ps 1  \\
      j & \shortminus1  &  \ps 1  &  \ps 0  \\
      k &  \ps 1  &  \ps 0  & \shortminus1  \\
  },
\end{equation}
and zeros everywhere else, so that the Frobenius inner product $\langle \bm{\mathcal{L}}_t, \bm{G}\rangle_F = \norm{\bm{y}_i - \bm{y}_k}_2^2 - \|\bm{y}_i - \bm{y}_j\|_2^2$ (and therefore, $w_t$ contributes only a sign). After minimizing \autoref{eq:empirical_risk}, we can recover $\bm{Y}$ from $\bm{G}$ up to a rigid transformation using the SVD.

In a maximum likelihood framework, $\ell$ is induced by our choice of $f$, assuming that the $w_t$ are independent. For example, if $f$ is the logistic function $f(x) = 1/(1+\exp(-x))$, the induced loss is the logistic loss $\ell(x) = \log(1+\exp(x))$ \citep{Jain2016}. This setup is equivalent to Stochastic Triplet Embeddings \citep{VanDerMaaten2012}, since the logistic loss and softmax are equivalent.

\citep{Jain2016} proves that the error $R(\hat{\bm{G}}) - R(\bm{G^*})$ (where $\bm{G^*}$ is the true underlying Gram matrix associated to $\bm{D}^*$) is bounded with high probability if $|\S| = \mathcal{O}(mn\log(n))$ and consequently, $\norm{\bm{\hat{D}} - \bm{D}^*}_F$ is also bounded. Therefore, the practical number of triplets that need to be queried is $\mathcal{O}(mn\log(n))$ instead of $\mathcal{O}(n^3)$.

When computing a 1-dimensional embedding (i.e., $m=1$), each $y_i\in\R$ can be interpreted as the value that the embedding takes at time index $i$, therefore representing a time series.
\section{Labeling triplets with multiple annotators}\label{sec:approach}
\autoref{eq:triplet_noise} shows a way to encode the decision of a single annotator when queried for a decision as in \autoref{eq:decision}. However, for multiple annotators we need to extend this model. Let $\mathcal{A}$ be a set of annotators. We define $\mathcal{S}_a$ as the set of triplets annotated by annotator $a\in\mathcal{A}$, so we observe a random variable $w_t^a$ for each $t\in\S_a$. The labels are defined as:
\begin{equation}
    w^a_t =
    \begin{cases}
      -1, &\text{w.p. } f_a(\bm{D}^*_{ij} - \bm{D}^*_{ik})\\
      +1, &\text{w.p. } 1 - f_a(\bm{D}^*_{ij} - \bm{D}^*_{ik}).
    \end{cases}
\end{equation}
where $f_a$ is the function that drives the probabilities for each annotator.

\subsection{Annotation fusion}
Due to annotation costs, we choose the sets $\mathcal{S}_a$ such that they are disjoint:
\begin{equation}
    \mathcal{S} = \bigcup_{a\in\mathcal{A}} \mathcal{S}_a \quad\text{ and }\quad \bigcap_{a\in\mathcal{A}}\mathcal{S}_a = \varnothing,
\label{eq:set_T_properties}
\end{equation}
so that all queries are unique and any annotated triplet $(i,j,k)$ is labeled by at most one annotator.

Note that the fusion process occurs in this step: The annotation fusion in a triplet embedding approach is done by taking the union of all the individually generated sets $\mathcal{S}_a$ to generate a single set of triplets $\mathcal{S}$, and using all corresponding labels $w_t^a$, defined for each annotator and each corresponding triplet $t\in\S$.

One difficulty of this multi-annotator model is that the distribution of $w_t^a$ depends on the annotators through $f_a$, and, hence, the loss function is annotator-dependent. Fortunately, in our experiments, we can assume $f_a=f$, as we show experimentally in \autoref{fig:MTurk_annotators}. We will extend this to annotator-dependent distributions in a future communication.

\subsection{Triplet violations and annotation agreements}
Triplet violations occur when a given triplet $t = (i,j,k)\in\S$ does not follow the calculated embedding $\bm{Y}$:
\begin{equation}
\|\bm{y}_i - \bm{y}_k\|_2 < \|\bm{y}_i - \bm{y}_j\|_2, \quad (i,j,k)\in\S.
\label{eq:triplet_violation}
\end{equation}
Therefore, we can count the fraction of triplet violations using:
\begin{equation}
    \tau_v = \frac{1}{|\S|}\sum_{(i,j,k)\in\mathcal{S}} \delta\left[\|\bm{y}_i - \bm{y}_k\|_2 < \|\bm{y}_i - \bm{y}_j\|_2\right]
    \text{,}
\end{equation}
where $\delta[\cdot]$ is Kronecker's delta.

To compute the expected number of correctly labeled triplets in $\S$, we can derive another random variable that models the correct annotation of triplet $t = (i,j,k)$ based on $f$:
\begin{equation}
    c_t =
    \begin{cases}
      0, &\text{w.p. } 1 - f(|\bm{D}^*_{ij} - \bm{D}^*_{ik}|)\\
      1, &\text{w.p. } f(|\bm{D}^*_{ij} - \bm{D}^*_{ik}|)\text{,}
    \end{cases}\label{eq:correct_annotation}
\end{equation}
where $f(|\bm{D}^*_{ij} - \bm{D}^*_{ik}|)$ is the probability of successfully annotating triplet $(i,j,k)$.

Using \autoref{eq:correct_annotation} we can model the number of correctly labeled triplets as a Poisson binomial random variable $C$:
\begin{equation}
C = \sum_{t=(i,j,k)\in\S} c_t \sim \text{PBD}\left(f(|\bm{D}^*_{ij} - \bm{D}^*_{ik}|), |\S|\right).
\end{equation}
Its expected value is the sum of the success probabilities:
\begin{equation}
\E[C] = \sum_{t=(i,j,k)\in\S} f(|\bm{D}^*_{ij} - \bm{D}^*_{ik}|). \label{eq:PBD-expectation}
\end{equation}

After computing $\bm{Y}$ from $\S$, and assuming that the optimization routine has found the best possible embedding $\bm{Y}$ for $\S$, then
the fraction of triplet violations $\tau_v$ in $\bm{Y}$ is linearly related to $C$ by:
\begin{equation}
	\tau_v = 1 - C/|\S|, \text{ or }\, \E[\tau_v] = 1 - \E[C]/|\S|.
\end{equation}
$\tau_v\in[0,1]$ is a measure of disagreement between all triplets used to compute the embedding $\bm{Y}$. $\tau_v = 0$ means that all used triplets agree with the computed embedding $\bm{Y}$, meaning that all triplet labels agree with each other. 
\section{Experiments}\label{sec:experiments}
We conduct two simulation experiments and one human annotation experiment using Mechanical Turk to verify the efficacy of our approach. We use the two synthetic data sets proposed in \citep{Booth2018}, for which the values for $\bm{Z}$ are known. We use this data because the reconstruction errors can be computed and we can assess the quality of the resulting labels, in contrast to experiments with affect, where the underlying signal is unknown. The two tasks correspond to videos of green frames with varying intensity of color over time and where the hidden construct $\bm{Z}$ is the intensity of green color (shown in thick black lines in \autoref{fig:example}). The video in task A is 267s long, and 178s long in task B.

To construct our triplet problem we first downsample the videos to 1Hz, so that the number of frames $n$ equals the length of the video in seconds to reduce the number of unique triplets. We also set the dimension $m$ to $1$, since we want to find a 1-dimensional embedding that represents the intensity of green color over time.

Our experiments are implemented in Julia v1.0 \citep{Bezanson2017}, and available at \url{www.github.com/kmundnic/PRL2019}.

\subsection{Synthetic triplet annotations}\label{sec:simulations}
We simulate the annotation procedure by comparing the scalar green intensity values of frames of the video using the absolute value of the difference between points. Hence, the dissimilarity for \autoref{eq:decision} is $d(z_i,z_j) = |z_i - z_j|$, where $i$ and $j$ are time indices.

We generate a list of noisy triplets $\S$ by randomly and uniformly selecting each triplet $(i,j,k)$ from the pool of all possible unique triplets. Each triplet $t=(i,j,k)$ is \textit{correctly} labeled by $w_t$ with probability $f(|\bm{D}^*_{ij} - \bm{D}^*_{ik}|)$.

We test eight different fractions of the total possible number of triplets $|\T|$ using logarithmic increments such that  $|\S| = \{0.0005,\ldots,0.1077\}|\T|$, which goes from $0.05\%$ to $10.77\%$ of $|\T|$. We use a logarithmic scale to have more resolution for smaller percentages of the total number of possible unique triplets. Note that for 267 frames (task A), the total number of unique triplets is 9,410,415. The queried triplets are randomly and uniformly sampled from all possible unique triplets, since there is no guarantee of better performance for active sampling algorithms in this problem \citep{Jamieson2015}.

We use various algorithms available in the literature to solve the triplet embedding problem: Stochastic Triplet Embeddings (STE) \citep{VanDerMaaten2012} (with $\sigma=1/\sqrt{2}$) and t-Student Stochastic Triplet Embeddings (tSTE) \citep{VanDerMaaten2012} (with $\alpha \in \{2,10\}$), Generalized Non-metric Multidimensional Scaling (GNMDS) \citep{Agarwal2007} (parameter-free) with hinge loss, and Crowd Kernel Learning (CKL) \citep{Tamuz2011CKL} (with $\mu\in\{2,10\}$).
We use gradient descent to optimize all the loss functions proposed by the algorithms.
Note that STE and GNMDS pose convex problems, while tSTE and CKL pose non-convex problems, and therefore we perform 30 different random starts for each set of parameters.

We now describe the three experimental settings we use to validate our approach.

\paragraph*{Simulation 1: Constant success probabilities}
We choose $f(|\bm{D}^*_{ij} - \bm{D}^*_{ik}|)$ to be approximately constant, such that the probability $f(|\bm{D}^*_{ij} - \bm{D}^*_{ik}|) = \mu + \epsilon$, where $\epsilon\sim\mathcal{N}(0,\sigma^2)$, $\sigma=0.01$ (to add small variations). We run three different experiments for $\mu\in\{0.7,0.8,0.9\}$.

Picking the values of $f(|\bm{D}^*_{ij} - \bm{D}^*_{ik}|)$ randomly affects our calculation of $\E[C]$ (\autoref{eq:PBD-expectation}), but we will assume that these have been fixed \emph{a priori}, meaning that the annotation process has a fixed probability for labeling any triplet $(i,j,k)$.

\paragraph*{Simulation 2: Logistic probabilities}
A more realistic simulation is given by labeling the triplets in $\S$ according to the following probabilities:
\begin{equation}
    f(\bm{D}^*_{ij} - \bm{D}^*_{ik}) = \frac{1}{1 + \exp\left(-\sigma(\bm{D}^*_{ij} - \bm{D}^*_{ik})\right)},
\end{equation}
which is the logistic function. We use different values for $\sigma = \{2,6,20\}$. Intuitively, the triplets with smaller differences between $\bm{D}^*_{ij}$ and $\bm{D}^*_{ik}$ should be harder to label, and a more realistic noise model than constant errors independent of the difficulty of the task. Note that this noise model induces the logistic loss used in STE.

\subsection{Mechanical Turk triplet annotations}
Using the list of images generated earlier we sample $0.5\%$ of the total number of triplets of images randomly and uniformly. In this setting, we sample approximately $Kn\log(n)$ triplets, with $K = 31.5$ for task A, and $K = 15$ for task B. To compute the embedding we use STE with parameter $\sigma = 1/\sqrt{2}$.

\begin{figure}[t!]
    \centering
    \includegraphics{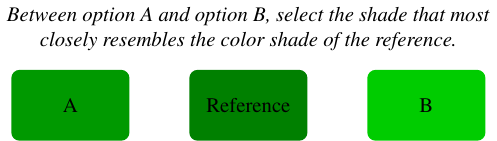}
    \caption{Question design for queries in Mechanical Turk.}
    \label{fig:MTurk_question_design}
    \vspace{-0.4cm}
\end{figure}

To obtain the list of annotated triplets, we show the annotators options A and B against a reference, and instructions as in \autoref{fig:MTurk_question_design}. We do not provide further instructions for the case where $\bm{D}^*_{Reference, A} \approx \bm{D}^*_{Reference, B}$.
For this task, we paid the annotators \$0.02 per answered query.

\begin{figure*}[t!]
    \centering
    \includegraphics{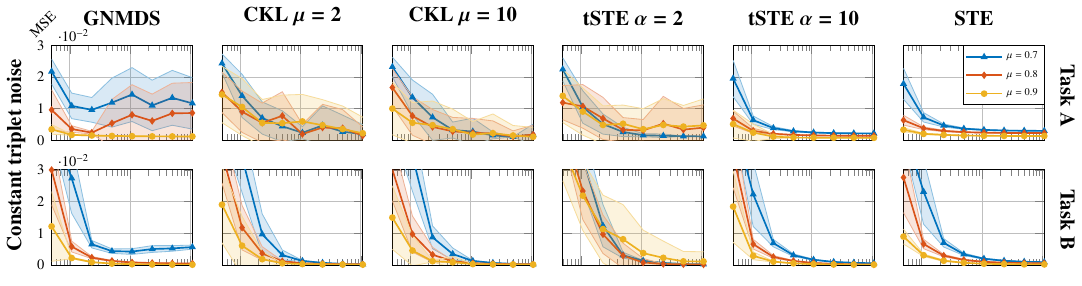}
    \includegraphics{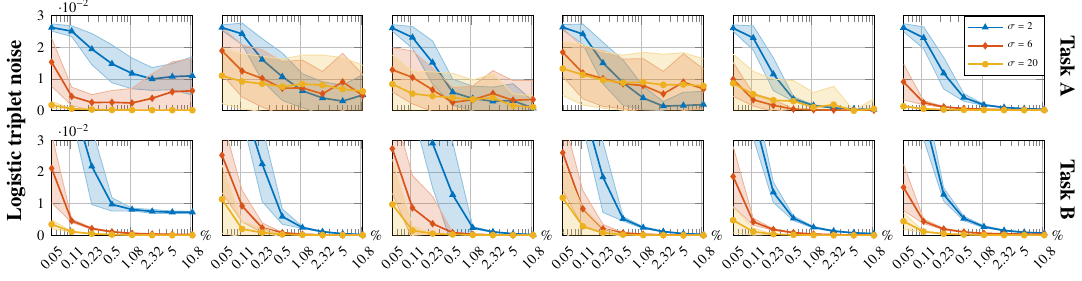}
    \vspace{-0.4cm}
    \caption{MSE as a function of the number of observed triplets $|\S|$ with constant and logistic noise in triplet labels. Each point in the plots represents the mean over 30 random trials, while the shaded areas represent one standard deviation from the average MSE values.}
    \vspace{-0.4cm}
    \label{Fig:MSE_noise}
\end{figure*}

\subsection{Error measure}
We use the error measure proposed in \citep{Terada2014}, and compute the error by first solving the following optimization problem:
\begin{equation}
	\text{MSE} = \inf_{a,b} \frac{1}{n}\norm{a\bm{Y} - b\bm{1} - \bm{Z}}^2_2, \label{eq:MSE}
\end{equation}
where $a, b \in \R$ are the scaling and bias factors, and $n$ is the length of $\bm{Y}$. We use this MSE and not a naive MSE between the ground truth $\bm{Z}$ and the reconstructed label $\bm{Y}$ because the embeddings are optimal only up to scaling and bias factors. Hence, this approach yields a more fair assessment of the quality of the embedding.

We also report Pearson's correlation $\rho$ between the ground truth and the estimated embedding, to compare our method with other proposed algorithms in a scale-free manner.

\subsection{Comparison to other methods}
We compare the proposed annotation and fusion framework with two different approaches using real-time annotations: EvalDep \cite{Mariooryad2015} and the EM-based approach (after time-alignment using EvalDep's method) from \cite{gupta2016modeling} with window lengths of 4, 8, 16, and 32.

\section{Results and analysis}\label{sec:analysis}
\subsection{Synthetic annotations}
\autoref{Fig:MSE_noise} shows the MSEs as a function of $|\S|/|\T|\times 100$ for both synthetic experiments. For both constant and logistic noise in tasks A and B we generally obtain a better performance as the amount of noise in the triplet annotation process is reduced (larger $\mu$ or $\sigma$). This is not always true in the algorithms that propose non-convex loss functions (tSTE, CKL), where sometimes more noise generates better embeddings. We \textit{hypothesize} that these algorithms sometimes find better local minima under noisier conditions.

The MSE in \autoref{Fig:MSE_noise} typically becomes smaller as $|\S|$ increases. This is true (generally) for tSTE, STE, and CKL. GNMDS does not always produce a better embedding by increasing the number of triplets employed.

We also note that the embedding in task B is easier to compute than that of Task A. We observe two possible reasons for this: (1) Task A has constant intervals while task B has none (and constant regions may be harder to compute in noisy conditions), and (2) the extreme values in task A seem harder to estimate, since these occur for very short intervals of time that are less likely to be sampled.

Overall, STE is the best-performing algorithm independent of noise or task. We note that tSTE with $\alpha = 10$ \textit{approaches} STE in many of the presented scenarios. In fact, tSTE becomes STE with $\sigma \to 1$ as $\alpha \to \infty$, so these results are expected (we do not include the proof due to space restrictions).

\subsection{Mechanical Turk triplet annotations}
\subsubsection{Annotator noise}
In the Mechanical Turk experiments, 170 annotators annotated triplets in task A, and 153 in task B. To understand the difficulty of the tasks and the noise distributions for the annotators, we estimate the probabilities of success $f(|\bm{D}^*_{ij} - \bm{D}^*_{ik}|)$ for both tasks, using the top three annotators.

To estimate $f(|\bm{D}^*_{ij} - \bm{D}^*_{ik}|)$, we partition the triplets based on $|\bm{D}^*_{ij} - \bm{D}^*_{ik}|$ into intervals with the same number of triplets. For each interval, we compute the average distance of the triplets. For each triplet $(i,j,k)\in\mathcal{I}$, we know the outcome (realization) of the random variable $w_{ijk}^a$ since we know the hidden construct $\bm{Z}$. We assume that the success probability $f_a(|\bm{D}^*_{ij} - \bm{D}^*_{ik}|)$ is constant in this interval, so that $C\sim \text{Binomial}(f_a(|\bm{D}^*_{ij} - \bm{D}^*_{ik}|))$. Finally, we use the maximum likelihood estimator for success probabilities for each interval:
\begin{equation}
    \hat{f}_{I^a}(|\bm{D}^*_{ij} - \bm{D}^*_{ik}|) = \frac{1}{|\mathcal{I}^a|}\sum_{(i,j,k)\in\mathcal{I}^a} c_{ijk}^a.
\end{equation}

\begin{figure}[t!]
    \centering
        \begin{subfigure}{\linewidth}
        \centering
        \includegraphics{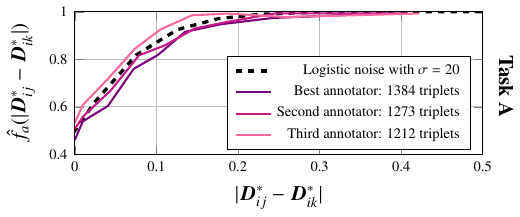}
    \end{subfigure}
    \begin{subfigure}{\linewidth}
        \centering
        \includegraphics{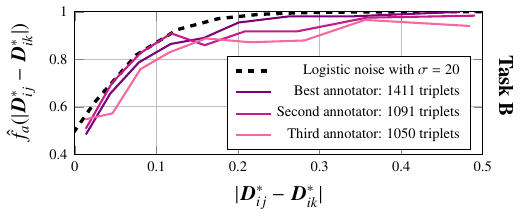}
    \end{subfigure}
    \caption{Probabilities of success $\hat{f}_{I^a}(|\bm{D}^*_{ij} - \bm{D}^*_{ik}|)$ as a function of the distance from the reference $i$ to frames $j$ and $k$. Only the top annotators have been included.}
    \label{fig:MTurk_annotators}
\end{figure}

In \autoref{fig:MTurk_annotators}, we show the function $\hat{f}_a(|\bm{D}^*_{ij} - \bm{D}^*_{ik}|)$ for each of the top annotators with the most answered queries and compare it to the logistic function with $\sigma=20$. The comparison between the estimated probabilities of success and the logistic function shows that this is a very good noise model for this annotation task, while also telling us that we should expect the best results from STE when computing the embedding from the crowd-sourced triplet annotations. \textit{Noticeably, our initial assumption of an annotator-independent noise model is verified}.

\subsubsection{Mechanical Turk embedding}
\begin{figure}[t!]
    \centering
        \begin{subfigure}{\linewidth}
        \centering
        \includegraphics{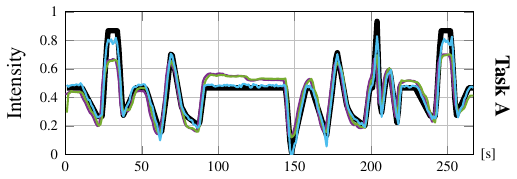}
    \end{subfigure}
    \begin{subfigure}{\linewidth}
        \centering
        \includegraphics{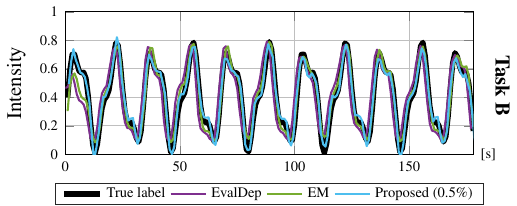}\hfill
    \end{subfigure}
    \caption{
    Results for Mechanical Turk annotations.
    The computed embeddings have been scaled to fit the true labels $\bm{Z}$ (\autoref{eq:MSE}).
    The embedding in task A uses 0.5\% (47,052) of all possible triplet comparisons $|\T|$.
    The embedding in task B uses 0.5\% (13,862) of all possible triplet comparisons $|\T|$.
    In both tasks, the estimated green intensity is sometimes less than zero due to scaling.}
    \label{fig:MTurk}
    \vspace{-0.4cm}
\end{figure}

We present in \autoref{fig:MTurk} the results for the reconstructed embeddings using triplets generated by annotators via Mechanical Turk. We show the reconstructed embeddings obtained using 0.5\% of the total number of triplets $|\T|$ for each task. Although there is some visible error, we are able to capture the trends and overall shape of the underlying construct with only 0.5\% or fewer of all possible triplets for both tasks. We also plot a scaled version (according to \autoref{eq:MSE}) of the fused annotation obtained using the EvalDep method and using the continuous-time annotations from \cite{Booth2018} (\autoref{fig:example}). In this figure, we observe that fusion methods based on continuous-time annotations are not able to debias the annotations in windows of time that are biased. Our proposed method does not suffer from this issue.

We show in \autoref{tab:mturk_results} the MSE for each task, where percentages again represent the number of triplets employed. We have also included the MSE and $\rho$ for the embedding produced with $0.25\%$ of the triplets (not included in \autoref{fig:MTurk} due to the high overlap between the $0.25\%$ and $0.5\%$ embeddings). We observe that the MSE is lower for a higher number of labeled triplets used. This is expected: there is more information about the embedding as we increase the number of triplets that we feed into the optimization routine, therefore producing a higher quality embedding. We also show a scale-free comparison through Pearson's correlation, which captures how signals vary over time and neglects differences in scale and bias. In task A, our approach improves upon previous work by a large margin. In task B, our approach performs comparably to the EM-based method. Our understanding suggests that the EM-based algorithm benefits from a smooth ground truth, given their filter-modeling approach on a given window size.

\subsubsection{Triplet violations and annotator agreement}
\autoref{tab:mturk_violations} displays the number of triplet violations for each task. We record the true percentage of triplet violations according to our ground truth (generated using distances $d(z_i,z_j)$ and $d(z_i,z_k)$, as in \autoref{eq:triplet_violation}) and then compare them to the annotation responses. We also display the number of triplet violations according to the computed embeddings $\bm{Y}$. We see that the percentage of triplet violations according to our ground truth and the triplet violations calculated from the embeddings $\bm{Y}$ is not the same, being overestimated in task A and underestimated in task B. We also observe that even if the number of violations increases in task A, the MSE is reduced with a larger number of triplets. This happens because a higher number of triplet constraints more easily define an embedding.

\begin{table}[t!]
\centering
\caption{MSE and Pearson's correlation $\rho$ for the proposed method and state of the art continuous-time fusion techniques against ground truth. For our method, percentage is with respect to the total number of triplets.}
\label{tab:mturk_results}
\scalebox{0.85}{
\begin{tabular}{clcc}
\toprule
 \textbf{Task} & \textbf{Fusion technique} & \textbf{MSE} & $\bm{\rho}$ \\
 \toprule
 \multirow{4}{*}{\textbf{A}}
 & EvalDep \cite{Mariooryad2015} & 0.00489 & 0.906 \\
 & EM \cite{gupta2016modeling} (best, window length: 16) & 0.00494 & 0.903 \\
 & Proposed (0.25\%) & 0.00145 & 0.973 \\
 & Proposed (0.50\%) & \textbf{0.00132} & \textbf{0.975} \\
 \midrule
 \multirow{4}{*}{\textbf{B}}
 & EvalDep \cite{Mariooryad2015} & 0.00304 & 0.969 \\
 & EM \citep{gupta2016modeling} (best, window length: 32) & \textbf{0.00241} & \textbf{0.975} \\
 & Proposed (0.25\%) & 0.00305 & 0.969 \\
 & Proposed (0.50\%) & 0.00285 & 0.971 \\
\bottomrule
\end{tabular}
}
\vspace{-0.5cm}
\end{table}

\begin{table}[t!]
    \centering
    \caption{Triplet violations $\tau_v$ for the Mechanical Turk experiment. Percentages correspond to percentage of total triplets observed. We include the fraction of triplet violations as computed by the labels generated with EvalDep.}
    \label{tab:mturk_violations}
    \scalebox{0.85}{
    \begin{tabular}{cccccc}
               \toprule
         & \multicolumn{4}{c}{\textbf{Triplet violations} $\bm{\tau_v}$} \\\cmidrule(lr){2-6}
         \textbf{Task} & MTurk   & $\bm{Y}$ (0.25\%) & $\bm{Y}$ (0.5\%) & EvalDep \cite{Mariooryad2015} & EM \cite{gupta2016modeling}\\\midrule
        \textbf{A}     & $0.122$ & $0.159$           & $0.161$        & 0.262                         & 0.259 \\
        \textbf{B}     & $0.179$ & $0.146$           & $0.129$        & 0.139                         & 0.124\\\bottomrule
    \end{tabular}
    }
\end{table}

\section{Discussion}\label{sec:discussion}
\autoref{sec:analysis} shows that it is possible to use triplet embeddings to find a 1-dimensional embedding that resembles the true underlying construct up to scaling and bias factors. There are several additional considerations for our proposed method.

\paragraph*{Annotation costs}
One of the challenging aspects of using triplet embeddings is the $\mathcal{O}(n^3)$ growth of the number of unique triplets for $n$ objects or frames. As mentioned earlier, the results by \citep{Jain2016} suggest however that the theoretical number of triplets needed scales with $\mathcal{O}(mn\log(n))$. In our experiments, we use $Kn\log(n)$ triplets with $K=31.5$ for task A and $K = 15$ for task B to achieve equivalent or better approximations of the underlying ground truth compared to the state-of-the-art.

\paragraph*{Computational costs}

Triplet embeddings are computationally cheap in comparison to the other methods employed in this paper, since they can be efficiently estimated using gradient-based methods to minimize the loss function. Moreover, in our current implementation, the calculation of the gradient is parallelized, and the average time to estimate the embeddings for task A and B using 0.5\% of $|\T|$ in \autoref{fig:MTurk} is 125 ms and 44 ms respectively over 100 trials using 10 threads on a laptop with an Intel i7-8850H processor and 32Gb of RAM.

The evaluation of the gradient for a fixed dimension $m$ of the embedding scales linearly with the number of triplets employed (where the number of triplets needed is $\mathcal{O}(n\log n)$. This is computationally cheap in comparison to other methods considered: for example, time-alignment (needed in all continuous-time fusion approaches) is an expensive operation. For a signal of length $n$, alignment using Dynamic Time Warping is $\mathcal{O}(n^2)$, and EvalDep \cite{Mariooryad2013, Mariooryad2015} needs to compute the determinants of two $n\times n$ matrices and one $2n\times 2n$ matrix to estimate the mutual information, each being at least $\mathcal{O}(n^{2.373})$ (by using Fast Matrix Multiplication \cite{aho1974}). The EM-based approach also requires inverting an $n\times n$ matrix. As examples, EvalDep takes 8s in task A and 6s in task B, while EM takes on average 20min for both tasks and different window lengths.

\paragraph*{Embedding quality}
The embeddings reconstructed are more accurate than the method proposed in \citep{Mariooryad2015}. Moreover, no time-alignment is needed since the annotation process does not suffer from reaction times. It is also important to note that sharp edges (high frequency regions of the construct) are most appropriately represented and do not get smoothed out, as with averaging-based annotation fusion techniques (where annotation devices such as mice or joysticks and user interfaces perform low-pass filtering).

In terms of reconstruction, the scaling factor is an open challenge. We see two possible ways to work with the differences in scaling when the underlying construct is unknown: (1) Learn the scaling in a machine learning pipeline that uses these labels to create a statistical model of the hidden construct, or (2) normalize the embedding $\bm{Y}$ such that $\bar{\bm{Y}} = 0$ and $\sigma_{\bm{Y}} = 1$, and train the models using either these labels or the derivatives $d\bm{Y}/dt$. However, we note that continuous-time annotations do suffer from the same loss of scaling and bias, since both techniques are trying to solve an inverse problem where the scale is not accessible.

\paragraph*{Feature sub-sampling for triplet comparisons}
In the experiments of this paper, we sub-sample the videos to 1Hz so that we have a manageable number of frames $n$. Down-sampling is possible due to the nature of the synthetic experiment we have created, but may not be suitable for other constructs such as affect in real world data, where annotation of single frames might lose important contextual information. In these scenarios, further investigation is needed to understand how to properly sub-sample more complex annotation tasks. \vspace{-0.2cm}

\section{Conclusion}\label{sec:conclusion}
In this paper, we present a new sampling methodology based on triplet comparisons to produce continuous-time labels of hidden constructs. To study the proposed methodology, we use two experiments previously proposed in \cite{Booth2018} and show that it is possible to recover the structure of the underlying hidden signals in simulation studies using human annotators to perform the triplet comparisons. These labels for the hidden signals are accurate up to scaling and bias factors.

Our method performs annotator fusion seamlessly as a union of sets of queried triplets $\S_a$, which greatly simplifies the fusion approach compared to existing approaches which directly combine real-time signals. Moreover, our approach does not need post-processing such as time-alignments or averaging.

Some challenges for the proposed method include dealing with the annotation costs given the number of triplets that needs to be sampled, and also learning the unknown scaling and bias factors.

As future directions, we are interested in several paths. We believe it is necessary to further study the proposed method for labeling constructs where the the ground truth cannot be validated, as is the case of human emotions, and contrast the effects of using triplet comparisons to annotate individual frames and using triplet comparisons to annotate over frame sequences.

\section*{Acknowledgements}
This work was supported by the National Science Foundation grant number 151454.

We thank Anil Ramakrishna for sharing with us the code for the EM-based approach.

\bibliographystyle{model2-names}

\begin{thebibliography}{31}
\expandafter\ifx\csname natexlab\endcsname\relax\def\natexlab#1{#1}\fi
\providecommand{\url}[1]{\texttt{#1}}
\providecommand{\href}[2]{#2}
\providecommand{\path}[1]{#1}
\providecommand{\DOIprefix}{doi:}
\providecommand{\ArXivprefix}{arXiv:}
\providecommand{\URLprefix}{URL: }
\providecommand{\Pubmedprefix}{pmid:}
\providecommand{\doi}[1]{\href{http://dx.doi.org/#1}{\path{#1}}}
\providecommand{\Pubmed}[1]{\href{pmid:#1}{\path{#1}}}
\providecommand{\bibinfo}[2]{#2}
\ifx\xfnm\relax \def\xfnm[#1]{\unskip,\space#1}\fi
\bibitem[{Agarwal et~al.(2007)Agarwal, Wills, Cayton, Lanckriet, Kriegman and
  Belongie}]{Agarwal2007}
\bibinfo{author}{Agarwal, S.}, \bibinfo{author}{Wills, J.},
  \bibinfo{author}{Cayton, L.}, \bibinfo{author}{Lanckriet, G.},
  \bibinfo{author}{Kriegman, D.}, \bibinfo{author}{Belongie, S.},
  \bibinfo{year}{2007}.
\newblock \bibinfo{title}{{Generalized Non-Metric Multidimensional Scaling}},
  in: \bibinfo{booktitle}{Artificial Intelligence and Statistics}, pp.
  \bibinfo{pages}{11--18}.
\bibitem[{Aho et~al.(1974)Aho, Hopcroft and Ullman}]{aho1974}
\bibinfo{author}{Aho, A.V.}, \bibinfo{author}{Hopcroft, J.E.},
  \bibinfo{author}{Ullman, J.D.}, \bibinfo{year}{1974}.
\newblock \bibinfo{title}{{The Design and Analysis of Computer Algorithms}}.
\newblock \bibinfo{publisher}{Addison-Wesley}.
\bibitem[{Bezanson et~al.(2017)Bezanson, Edelman, Karpinski and
  Shah}]{Bezanson2017}
\bibinfo{author}{Bezanson, J.}, \bibinfo{author}{Edelman, A.},
  \bibinfo{author}{Karpinski, S.}, \bibinfo{author}{Shah, V.B.},
  \bibinfo{year}{2017}.
\newblock \bibinfo{title}{{Julia: A Fresh Approach to Numerical Computing}}.
\newblock \bibinfo{journal}{SIAM Review} \bibinfo{volume}{59},
  \bibinfo{pages}{65--98}.
\bibitem[{Booth et~al.(2018a)Booth, Mundnich and Narayanan}]{Booth2018}
\bibinfo{author}{Booth, B.M.}, \bibinfo{author}{Mundnich, K.},
  \bibinfo{author}{Narayanan, S.}, \bibinfo{year}{2018}a.
\newblock \bibinfo{title}{{A Novel Method for Human Bias Correction of
  Continuous-Time Annotations}}, in: \bibinfo{booktitle}{2018 IEEE
  International Conference on Acoustics, Speech and Signal Processing
  (ICASSP)}, \bibinfo{organization}{IEEE}. pp. \bibinfo{pages}{3091--3095}.
\bibitem[{Booth et~al.(2018b)Booth, Mundnich and Narayanan}]{Booth2018AVEC}
\bibinfo{author}{Booth, B.M.}, \bibinfo{author}{Mundnich, K.},
  \bibinfo{author}{Narayanan, S.}, \bibinfo{year}{2018}b.
\newblock \bibinfo{title}{{Fusing Annotations with Majority Vote Triplet
  Embeddings}}, in: \bibinfo{booktitle}{Proceedings of the 2018 Audio/Visual
  Emotion Challenge and Workshop}, \bibinfo{organization}{ACM}. pp.
  \bibinfo{pages}{83--89}.
\bibitem[{Busso et~al.(2013)Busso, Bulut and Narayanan}]{Busso2013}
\bibinfo{author}{Busso, C.}, \bibinfo{author}{Bulut, M.},
  \bibinfo{author}{Narayanan, S.}, \bibinfo{year}{2013}.
\newblock \bibinfo{title}{{Toward Effective Automatic Recognition Systems of
  Emotion in Speech}}.
\newblock \bibinfo{journal}{Social emotions in nature and artifact: emotions in
  human and human-computer interaction, J. Gratch and S. Marsella, Eds} ,
  \bibinfo{pages}{110--127}.
\bibitem[{Cowie and Cornelius(2003)}]{Cowie2003}
\bibinfo{author}{Cowie, R.}, \bibinfo{author}{Cornelius, R.R.},
  \bibinfo{year}{2003}.
\newblock \bibinfo{title}{{Describing the Emotional States that are Expressed
  in Speech}}.
\newblock \bibinfo{journal}{Speech Communication} \bibinfo{volume}{40},
  \bibinfo{pages}{5--32}.
\bibitem[{Davenport et~al.(2014)Davenport, Plan, Van Den~Berg and
  Wootters}]{davenport2014}
\bibinfo{author}{Davenport, M.A.}, \bibinfo{author}{Plan, Y.},
  \bibinfo{author}{Van Den~Berg, E.}, \bibinfo{author}{Wootters, M.},
  \bibinfo{year}{2014}.
\newblock \bibinfo{title}{{1-bit Matrix Completion}}.
\newblock \bibinfo{journal}{Information and Inference: A Journal of the IMA}
  \bibinfo{volume}{3}, \bibinfo{pages}{189--223}.
\bibitem[{Ellis et~al.(2002)Ellis, Whitman, Berenzweig and
  Lawrence}]{ellis2002quest}
\bibinfo{author}{Ellis, D.P.W.}, \bibinfo{author}{Whitman, B.},
  \bibinfo{author}{Berenzweig, A.}, \bibinfo{author}{Lawrence, S.},
  \bibinfo{year}{2002}.
\newblock \bibinfo{title}{{The Quest for Ground Truth in Musical Artist
  Similarity}}, in: \bibinfo{booktitle}{In Proceedings of the International
  Symposium on Music Information Retrieval (ISMIR 2002)}, pp.
  \bibinfo{pages}{170--177}.
\bibitem[{Gupta et~al.(2018)Gupta, Audhkhasi, Jacokes, Rozga and
  Narayanan}]{gupta2016modeling}
\bibinfo{author}{Gupta, R.}, \bibinfo{author}{Audhkhasi, K.},
  \bibinfo{author}{Jacokes, Z.}, \bibinfo{author}{Rozga, A.},
  \bibinfo{author}{Narayanan, S.}, \bibinfo{year}{2018}.
\newblock \bibinfo{title}{{Modeling Multiple Time Series Annotations as Noisy
  Distortions of the Ground Truth: An Expectation-Maximization Approach}}.
\newblock \bibinfo{journal}{IEEE Transactions on Affective Computing}
  \bibinfo{volume}{9}, \bibinfo{pages}{76--89}.
\newblock \DOIprefix\doi{10.1109/TAFFC.2016.2592918}.
\bibitem[{Hotelling(1936)}]{Hotelling1936}
\bibinfo{author}{Hotelling, H.}, \bibinfo{year}{1936}.
\newblock \bibinfo{title}{{Relations between Two Sets of Variates}}.
\newblock \bibinfo{journal}{Biometrika} \bibinfo{volume}{28},
  \bibinfo{pages}{321--377}.
\bibitem[{Jain et~al.(2016)Jain, Jamieson and Nowak}]{Jain2016}
\bibinfo{author}{Jain, L.}, \bibinfo{author}{Jamieson, K.G.},
  \bibinfo{author}{Nowak, R.D.}, \bibinfo{year}{2016}.
\newblock \bibinfo{title}{{Finite Sample Prediction and Recovery Bounds for
  Ordinal Embedding}}, in: \bibinfo{booktitle}{Advances in Neural Information
  Processing Systems}, pp. \bibinfo{pages}{2711--2719}.
\bibitem[{Jamieson et~al.(2015)Jamieson, Jain, Fernandez, Glattard and
  Nowak}]{Jamieson2015}
\bibinfo{author}{Jamieson, K.G.}, \bibinfo{author}{Jain, L.},
  \bibinfo{author}{Fernandez, C.}, \bibinfo{author}{Glattard, N.J.},
  \bibinfo{author}{Nowak, R.D.}, \bibinfo{year}{2015}.
\newblock \bibinfo{title}{{NEXT: A System for Real-World Development,
  Evaluation, and Application of Active Learning}}, in:
  \bibinfo{booktitle}{Advances in Neural Information Processing Systems}, pp.
  \bibinfo{pages}{2656--2664}.
\bibitem[{Mariooryad and Busso(2013)}]{Mariooryad2013}
\bibinfo{author}{Mariooryad, S.}, \bibinfo{author}{Busso, C.},
  \bibinfo{year}{2013}.
\newblock \bibinfo{title}{{Analysis and Compensation of the Reaction Lag of
  Evaluators in Continuous Emotional Annotations}}, in:
  \bibinfo{booktitle}{Affective Computing and Intelligent Interaction (ACII),
  2013 Humaine Association Conference on}, \bibinfo{organization}{IEEE}. pp.
  \bibinfo{pages}{85--90}.
\bibitem[{Mariooryad and Busso(2015)}]{Mariooryad2015}
\bibinfo{author}{Mariooryad, S.}, \bibinfo{author}{Busso, C.},
  \bibinfo{year}{2015}.
\newblock \bibinfo{title}{{Correcting Time-Continuous Emotional Labels by
  Modeling the Reaction Lag of Evaluators}}.
\newblock \bibinfo{journal}{IEEE Transactions on Affective Computing}
  \bibinfo{volume}{6}, \bibinfo{pages}{97--108}.
\bibitem[{McFee et~al.(2012)McFee, Barrington and
  Lanckriet}]{mcfee2012learning}
\bibinfo{author}{McFee, B.}, \bibinfo{author}{Barrington, L.},
  \bibinfo{author}{Lanckriet, G.}, \bibinfo{year}{2012}.
\newblock \bibinfo{title}{{Learning Content Similarity for Music
  Recommendation}}.
\newblock \bibinfo{journal}{IEEE transactions on audio, speech, and language
  processing} \bibinfo{volume}{20}, \bibinfo{pages}{2207--2218}.
\bibitem[{McFee and Lanckriet(2011)}]{mcfee2011learning}
\bibinfo{author}{McFee, B.}, \bibinfo{author}{Lanckriet, G.},
  \bibinfo{year}{2011}.
\newblock \bibinfo{title}{{Learning Multi-modal Similarity}}.
\newblock \bibinfo{journal}{Journal of machine learning research}
  \bibinfo{volume}{12}, \bibinfo{pages}{491--523}.
\bibitem[{Metallinou and Narayanan(2013)}]{Metallinou2013}
\bibinfo{author}{Metallinou, A.}, \bibinfo{author}{Narayanan, S.},
  \bibinfo{year}{2013}.
\newblock \bibinfo{title}{{Annotation and Processing of Continuous Emotional
  Attributes: Challenges and Opportunities}}, in: \bibinfo{booktitle}{Automatic
  Face and Gesture Recognition (FG), 2013 10th IEEE International Conference
  and Workshops on}, \bibinfo{organization}{IEEE}. pp. \bibinfo{pages}{1--8}.
\bibitem[{M{\"u}ller(2007)}]{DTW2007}
\bibinfo{author}{M{\"u}ller, M.}, \bibinfo{year}{2007}.
\newblock \bibinfo{title}{{Dynamic Time Warping}}.
\newblock \bibinfo{journal}{Information retrieval for music and motion} ,
  \bibinfo{pages}{69--84}.
\bibitem[{{Neil Stewart and Gordon D. A. Brown and Nick
  Chater}(2005)}]{Stewart2005}
\bibinfo{author}{{Neil Stewart and Gordon D. A. Brown and Nick Chater}},
  \bibinfo{year}{2005}.
\newblock \bibinfo{title}{{Absolute Identification by Relative Judgement}}.
\newblock \bibinfo{journal}{Psychological Review} \bibinfo{volume}{112},
  \bibinfo{pages}{881--911}.
\bibitem[{Nicolaou et~al.(2013)Nicolaou, Zafeiriou and Pantic}]{Nicolaou2013}
\bibinfo{author}{Nicolaou, M.A.}, \bibinfo{author}{Zafeiriou, S.},
  \bibinfo{author}{Pantic, M.}, \bibinfo{year}{2013}.
\newblock \bibinfo{title}{{Correlated-Spaces Regression for Learning Continuous
  Emotion Dimensions}}, in: \bibinfo{booktitle}{Proceedings of the 21st ACM
  international conference on Multimedia}, \bibinfo{organization}{ACM}. pp.
  \bibinfo{pages}{773--776}.
\bibitem[{Raykar et~al.(2010)Raykar, Yu, Zhao, Valadez, Florin, Bogoni and
  Moy}]{Raykar2010}
\bibinfo{author}{Raykar, V.C.}, \bibinfo{author}{Yu, S.},
  \bibinfo{author}{Zhao, L.H.}, \bibinfo{author}{Valadez, G.H.},
  \bibinfo{author}{Florin, C.}, \bibinfo{author}{Bogoni, L.},
  \bibinfo{author}{Moy, L.}, \bibinfo{year}{2010}.
\newblock \bibinfo{title}{{Learning from Crowds}}.
\newblock \bibinfo{journal}{Journal of Machine Learning Research}
  \bibinfo{volume}{11}, \bibinfo{pages}{1297--1322}.
\bibitem[{Ringeval et~al.(2015)Ringeval, Eyben, Kroupi, Yuce, Thiran, Ebrahimi,
  Lalanne and Schuller}]{Ringeval2015}
\bibinfo{author}{Ringeval, F.}, \bibinfo{author}{Eyben, F.},
  \bibinfo{author}{Kroupi, E.}, \bibinfo{author}{Yuce, A.},
  \bibinfo{author}{Thiran, J.P.}, \bibinfo{author}{Ebrahimi, T.},
  \bibinfo{author}{Lalanne, D.}, \bibinfo{author}{Schuller, B.W.},
  \bibinfo{year}{2015}.
\newblock \bibinfo{title}{{Prediction of Asynchronous Dimensional Emotion
  Ratings from Audiovisual and Physiological Data}}.
\newblock \bibinfo{journal}{Pattern Recognition Letters} \bibinfo{volume}{66},
  \bibinfo{pages}{22--30}.
\bibitem[{Tamuz et~al.(2011)Tamuz, Liu, Belongie, Shamir and
  Kalai}]{Tamuz2011CKL}
\bibinfo{author}{Tamuz, O.}, \bibinfo{author}{Liu, C.},
  \bibinfo{author}{Belongie, S.}, \bibinfo{author}{Shamir, O.},
  \bibinfo{author}{Kalai, A.T.}, \bibinfo{year}{2011}.
\newblock \bibinfo{title}{{Adaptively Learning the Crowd Kernel}}, in:
  \bibinfo{booktitle}{Proceedings of the 28th International Conference on
  Machine Learning}, pp. \bibinfo{pages}{673--680}.
\bibitem[{Terada and von Luxburg(2014)}]{Terada2014}
\bibinfo{author}{Terada, Y.}, \bibinfo{author}{von Luxburg, U.},
  \bibinfo{year}{2014}.
\newblock \bibinfo{title}{{Local ordinal embedding}}, in:
  \bibinfo{booktitle}{International Conference on Machine Learning}, pp.
  \bibinfo{pages}{847--855}.
\bibitem[{Trigeorgis et~al.(2018)Trigeorgis, Nicolaou, Schuller and
  Zafeiriou}]{trigeorgis2018deep}
\bibinfo{author}{Trigeorgis, G.}, \bibinfo{author}{Nicolaou, M.A.},
  \bibinfo{author}{Schuller, B.W.}, \bibinfo{author}{Zafeiriou, S.},
  \bibinfo{year}{2018}.
\newblock \bibinfo{title}{{Deep Canonical Time Warping for Simultaneous
  Alignment and Representation Learning of Sequences}}.
\newblock \bibinfo{journal}{IEEE Transactions on Pattern Analysis \& Machine
  Intelligence} \bibinfo{volume}{40}, \bibinfo{pages}{1128--1138}.
\bibitem[{Van Der~Maaten and Weinberger(2012)}]{VanDerMaaten2012}
\bibinfo{author}{Van Der~Maaten, L.}, \bibinfo{author}{Weinberger, K.},
  \bibinfo{year}{2012}.
\newblock \bibinfo{title}{{Stochastic Triplet Embedding}}, in:
  \bibinfo{booktitle}{Machine Learning for Signal Processing (MLSP), 2012 IEEE
  International Workshop on}, \bibinfo{organization}{IEEE}. pp.
  \bibinfo{pages}{1--6}.
\bibitem[{Yannakakis and Hallam(2011)}]{Yannakakis2011}
\bibinfo{author}{Yannakakis, G.N.}, \bibinfo{author}{Hallam, J.},
  \bibinfo{year}{2011}.
\newblock \bibinfo{title}{{Ranking vs. Preference: A Comparative Study of
  Self-Reporting}}.
\newblock \bibinfo{journal}{Affective Computing and Intelligent Interaction} ,
  \bibinfo{pages}{437--446}.
\bibitem[{Yannakakis and Mart{\'i}nez(2015)}]{Yannakakis2015}
\bibinfo{author}{Yannakakis, G.N.}, \bibinfo{author}{Mart{\'i}nez, H.P.},
  \bibinfo{year}{2015}.
\newblock \bibinfo{title}{{Ratings are Overrated!}}
\newblock \bibinfo{journal}{Frontiers in ICT} \bibinfo{volume}{2},
  \bibinfo{pages}{13}.
\bibitem[{Zhou and De~la Torre(2009)}]{Zhou2009}
\bibinfo{author}{Zhou, F.}, \bibinfo{author}{De~la Torre, F.},
  \bibinfo{year}{2009}.
\newblock \bibinfo{title}{{Canonical Time Warping for Alignment of Human
  Behavior}}, in: \bibinfo{booktitle}{Advances in neural information processing
  systems}, pp. \bibinfo{pages}{2286--2294}.
\bibitem[{Zhou and De~la Torre(2012)}]{zhou2012generalized}
\bibinfo{author}{Zhou, F.}, \bibinfo{author}{De~la Torre, F.},
  \bibinfo{year}{2012}.
\newblock \bibinfo{title}{{Generalized Time Warping for Multi-modal Alignment
  of Human Motion}}, in: \bibinfo{booktitle}{Computer Vision and Pattern
  Recognition (CVPR), 2012 IEEE Conference on}, pp.
  \bibinfo{pages}{1282--1289}.

\end{thebibliography}

\end{document}